\documentclass[journal,twoside,web]{ieeecolor}
\usepackage{tmi}
\usepackage{cite}
\usepackage{amsmath,amssymb,amsfonts}
\usepackage{algorithmic}
\usepackage{graphicx}
\usepackage{textcomp}
\usepackage{multirow}
\usepackage{subfig}

\def\BibTeX{{\rm B\kern-.05em{\sc i\kern-.025em b}\kern-.08em
    T\kern-.1667em\lower.7ex\hbox{E}\kern-.125emX}}
\begin{document}
\title{StainPresetNet: Stain Preset Network for Fast Multi-to-Multi Stain Normalization}
\author{Hongtao Kang, Die Luo, Li Chen, Jing Cai, Junbo Hu, Xiuli Liu, Shenghua Cheng
\thanks{This work is supported by the National Natural Science Foundation of China (grant 62201221, 62375100, and 62471212), China Postdoctoral Science Foundation (grant 2021M701320, 2022T150237), and Science and Technology Projects in Guangzhou (grant 2024A04J4960). \textit{(Corresponding author: Xiuli Liu and Shenghua Cheng.)}}
\thanks{H Kang and S Cheng are with the School of Biomedical Engineering and Guangdong Provincial Key Laboratory of Medical Image Processing, Southern Medical University, Guangzhou, Guangdong 510515, China(e-mail: khtao@smu.edu.cn; chengsh2023@smu.edu.cn).}
\thanks{X Liu is with the Britton Chance Center for Biomedical Photonics, Wuhan National Laboratory for Optoelectronics, Huazhong University of Science and Technology, Wuhan, Hubei 430074, China, and also with the MoE Key Laboratory for Biomedical Photonics, School of Engineering Sciences, Huazhong University of Science and Technology, Wuhan, Hubei 430074, China(e-mail:xlliu@mail.hust.edu.cn).}
\thanks{D Luo is with the School of Computer Science, Hubei University of Technology, Wuhan, Hubei 430068, China(email:luodie@hbut.edu.cn).}
\thanks{L Chen and J Cai are with the Department of Clinical Laboratory, Tongji Hospital, Huazhong University of Science and Technology, Wuhan, Hubei 430030,China(email:chenliisme@126.com; jingcai@hust.edu.cn).}
\thanks{J Hu is with the Department of Pathology, Hubei Maternal and Child Health Hospital, Wuhan, Hubei 430072, China(email:cqjbhu@163.com).}
}
\maketitle

\begin{abstract}
	
Stain normalization reduces color variations caused by variations in staining protocols and imaging conditions, thereby enhancing computer-aided diagnostic system performance. Traditional methods derive mapping relationships from individual or limited reference images through pixel-wise transformation, offering style flexibility but suffering from inaccurate color mapping extraction. While existing deep-learning-based approaches achieve accurate dataset-wide color mapping through complex neural networks, they face challenges including computational inefficiency, artifact generation, and fixed normalization directions requiring model retraining for directional changes. To address these limitations, we propose StainPresetNet - a novel framework that combines structural preservation with dataset-level color mapping while maintaining computational efficiency. Our method implements pixel-wise normalization guided by preset reference images, enabling multi-directional adaptability without retraining. Evaluations on cytopathology and histopathology datasets demonstrate that StainPresetNet achieves superior color mapping accuracy compared to conventional methods, effectively improves classifier generalization in diagnostic tasks, and reduces computational overhead by 90\% versus existing deep learning approaches. The proposed preset-guided mechanism facilitates flexible adjustment of normalization directions through simple reference image replacement, overcoming the directional rigidity of current deep-learning-based solutions.

\end{abstract}


\section{Introduction}
\label{sec:introduction}
\IEEEPARstart{D}{ue} to various factors, pathological images often exhibit significant color variations \cite{salvi2021impact}. The significant color variations in pathological images hinder the application of a unified algorithm for image analysis, thereby severely impacting the robustness of computer-aided diagnosis (CAD) systems \cite{salvi2021impact,roy2018study,tosta2019computational}. Stain normalization can standardize pathological images obtained from different sources and under varying conditions to a consistent color distribution, thereby achieving uniform or near-uniform staining effects \cite{roy2018study,tosta2019computational,salvi2021impact}. Moreover, stain normalization can adjust the background brightness, contrast of pathological images to a unified and appropriate range, thus enhancing image quality. This improvement not only benefits the performance of CAD systems but also aids pathologists in making more accurate and consistent diagnoses.

Although stain normalization methods have a rich research history, several issues need to be addressed before they can be considered truly practical \cite{tosta2019computational,roy2018study}. Conventional stain normalization methods, whether based on global matching or stain separation, rely on reference images to extract mapping relationships \cite{roy2018study}. However, these conventional methods struggle to extract accurate mapping relationships from a single reference image \cite{tosta2019computational}. 

As general methods, the deep learning-based staining normalization methods are not limited by pathological image types, staining methods, reagents, or imaging techniques, and can be applied to various datasets \cite{tosta2019computational}. Given that pathological images serve as critical evidence for disease diagnosis, stain normalization methods must objectively and faithfully preserve the information of the source images, making only color adjustments without altering the image content (such as introducing distortions or artifacts). Most deep learning-based stain normalization methods employ complex deep networks to perform the normalization \cite{zheng2020stain,kang2021stainnet}. Due to the complexity of deep networks and the instability of adversarial training, normalized images often suffer from distortions and artifacts, which is unacceptable in pathological images\cite{zheng2020stain,kang2021stainnet}. 

Although some researchers have tried to reduce anomalies by adding additional branches and losses\cite{zhou2019enhanced,mahapatra2020structure,shrivastava2021self}, these improvements often complicate the training process and fail to fundamentally eliminate anomalies. In addition, the normalization direction of existing deep learning-based methods is usually fixed, and when the normalization direction needs to be changed, retraining is needed to learn a new color-mapping relationship \cite{han2021dynamic,zhou2021decoupled}. However, training deep learning-based stain normalization methods is a complex and time-consuming task, as it usually involves manually adjusting dozens of hyperparameters, which may take several hours to days to complete.

To address these problems, we propose a fast multi-to-multi stain normalization network, StainPresetNet, and its training framework. Our method has the following features:

\textbf{1. Structure Preservation}: StainPresetNet performs normalization using a fully 1×1 convolutional sub-network, ensuring that only the colors are adjusted without altering the image’s texture or structure.

\textbf{2. Accurate Mapping Relationships}: StainPresetNet employs a backbone network that directly controls the fully 1×1 convolutional color mapping sub-network, enabling the network to handle complex color mapping relationships accurately.

\textbf{3. Multi-domain to Multi-domain Applicability}: StainPresetNet uses reference images to guide the direction of stain normalization. This allows it to normalize multiple different styles of images to a single style or transform a single image to multiple different styles by changing the reference image.

\textbf{4. High Computational Efficiency}: The backbone network in StainPresetNet runs at low resolution, and the color mapping sub-network has only 59 parameters, resulting in high computational efficiency.

\section{Related Works}
\subsection{Conventional Stain Normalization methods}
Conventional stain normalization methods can be categorized into two types: global normalization methods\cite{coltuc2006exact,reinhard2001color} and stain separation-based methods\cite{ruifrok2001quantification,khan2014nonlinear,alsubaie2017stain,macenko2009method,vahadane2016structure}.

Global normalization methods typically process images in the LAB color space and mainly include histogram specification methods \cite{coltuc2006exact} and color matching methods \cite{reinhard2001color}. Histogram specification achieves normalization by mapping the histogram of the source image to the histogram of the target image \cite{coltuc2006exact}. The color matching method normalizes the means and standard deviations of the source image's LAB color space to those of the target image \cite{reinhard2001color}. However, normalizing only the mean and standard deviation does not eliminate the impact of local differences within the images \cite{roy2018study}.

The principle behind stain separation-based methods is that the concentration of stains and their intensity in an image are linearly related in the Optical Density (OD) space, allowing for the separation of different stains into independent channels, which can then be normalized individually \cite{roy2018study}. These methods can be divided into supervised and unsupervised stain separation methods. Supervised stain separation methods use manual measurements or pre-trained classifiers to separate pixels stained by different stains, thereby obtaining a stain matrix \cite{ruifrok2001quantification,khan2014nonlinear}. Unsupervised stain normalization methods use techniques such as Independent Component Analysis \cite{alsubaie2017stain}, Non-Negative Matrix Factorization \cite{vahadane2016structure}, and Singular Value Decomposition \cite{macenko2009method} to extract the stain matrix.

\subsection{Deep Learning Based Methods}

Since aligned source and target images are difficult to obtain, the methods based on pix2pix gan often train by performing color transformation and reconstruction on the target image \cite{nishar2020histopathological,zhao2022restainnet,tellez2019quantifying,zanjani2018stain}. Common color transformations include grayscale \cite{zhao2022restainnet}, HSV \cite{tellez2019quantifying}, CIEL *a*b \cite{zanjani2018stain}, etc. In addition, some researchers have tried to directly use unaligned source and target images for training. These methods often add additional branches to constrain the content of the generated images, such as using diagnostic task losses or using pre-trained VGG19 \cite{simonyan2014very} to constrain the content and style of the generated images \cite{bentaieb2017adversarial}.

CycleGAN-based stain normalization methods use cycle consistency loss to constrain image content and are trained using real source and target images, allowing the learning of accurate mapping relationships. Shaban et al. used CycleGAN \cite{zhu2017unpaired} for stain normalization and named their method StainGAN \cite{shaban2019staingan}. Subsequently, Runz et al. and Lo et al. adopted similar approaches \cite{runz2021normalization}. Cai et al.replaced the original generator with a U-Net generator with residual structures \cite{cai2019stain}. Mahapatra et al. introduced self-supervised semantic guidance into CycleGAN training to reduce artifacts \cite{mahapatra2020structure}. Shrivastava et al. incorporated attention mechanisms and structural similarity loss \cite{shrivastava2021self}. 

CycleGAN\cite{zhu2017unpaired} is well-suited for transformation between two domains. However, for the problem of normalizing images from multiple domains to a single domain (multi-to-one transformation), it is uncertain to which specific domain the single domain images should be transformed, making the direction of mapping indeterminate. Although CycleGAN\cite{zhu2017unpaired} can be directly used by treating multiple different domains as a single domain, this approach is likely to result in performance loss. To apply CycleGAN\cite{zhu2017unpaired} to multi-to-one stain normalization tasks, researchers have proposed several different solutions. Zhou et al. suggested using a stain color matrix to control the mapping between the single domain and multiple domains \cite{zhou2019enhanced}. Nazki et al. proposed a solution similar to StarGAN \cite{choi2018stargan}, using a single generator for cycle consistency training \cite{nazki2022multipathgan}. 

StainNet \cite{kang2021stainnet} is a full 1×1 convolutional network. Due to network capacity limitations, it is only suitable for one-to-one color normalization . Based on StainNet, ParamNet \cite{kang2023paramnet} uses a backbone network to predict the weights and biases of the full 1×1 convolutional network, solving the problem of insufficient network capacity of StainNet. Our StainPresetNet proposed in this paper uses a reference image to guide the normalization direction on the basis of ParamNet, thereby having the ability to perform stain normalization transformation in multiple styles.

\section{Methods}
In this section, we introduce the network structure of ParamNet and its adversarial training framework.

\subsection{Network Structure of StainPresetNet}
\begin{figure*}[!ht]
	\centering
	\includegraphics[width=100 mm]{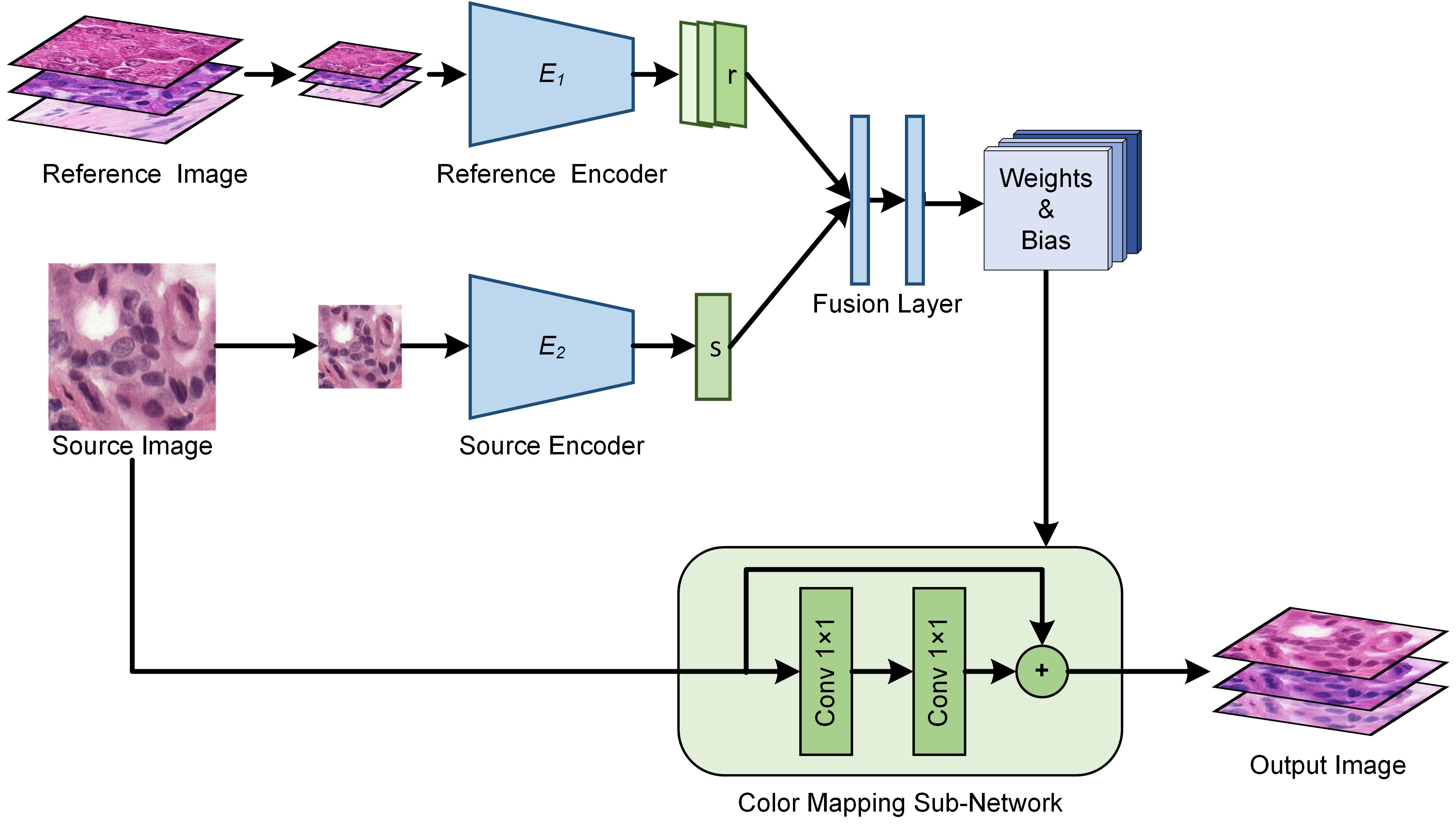}
	\caption{The network structure of StainPresetNet. The source encoder and reference encoder are used to extract color feature vectors, and the fusion layer is used to calculate the parameters of the color mapping sub-network, which is used to normalize the source image.}
	\label{fig1}
\end{figure*}

As illustrated in Fig. \ref{fig1}, StainPresetNet contains two encoders (a source encoder and a reference encoder), a fusion layer, and a color mapping sub-network. The source encoder extracts color distribution feature vectors from the source image, and the reference encoder extracts color distribution feature vectors from the reference image. The fusion layer calculates all parameters (weights and biases) of the color mapping sub-network based on the color distribution feature vectors from the source and reference images. The color mapping sub-network, a residual structure network with fully 1×1 convolutions, directly normalizes the source image. 

In our implementation, both the source encoder and the reference encoder use a modified ResNet18 \cite{he2016deep}, in which all batch normalization layers are removed and all convolutional layers are set as biased convolutional layers. The fusion layer contains two linear layers with the activation function of Leaky Rectified Linear Unit (LeakyReLU). For the stability of the network, the outputs of the fusion layer will be normalized by the tanh function, and then multiplied by a learnable parameter. The learnable parameter represents the numerical range of the parameters (weights and biases of convolutional layers) in the color mapping sub-network and its initial value is set to 4.5. The input images of the source encoder and the reference encoder are downsampled to 128×128 by default. 

During the running stage, the weights and biases of the color mapping sub-network are not fixed, but are determined by the style of the input image and the reference image, which enables our network to handle complex color mapping relationships. Meanwhile, the color mapping sub-network, which is a fully 1×1 convolutional network, guarantees structural consistency between the input and output images. The network design follows the principle of conducting complex computations at low resolutions and simple computations at high resolutions. Specifically, the source encoder, reference encoder, and fusion layer operate at low resolution, but the color mapping sub-network operates at the original resolution. During the running stage, the reference image is pre-set, and the color distribution feature vector of the reference image does not need to be repeatedly computed. These features endow StainPresetNet with high computational efficiency.

\subsection{Training Framework of StainPresetNet}

\begin{figure}[!ht]
	\centering
	\includegraphics[width=80 mm]{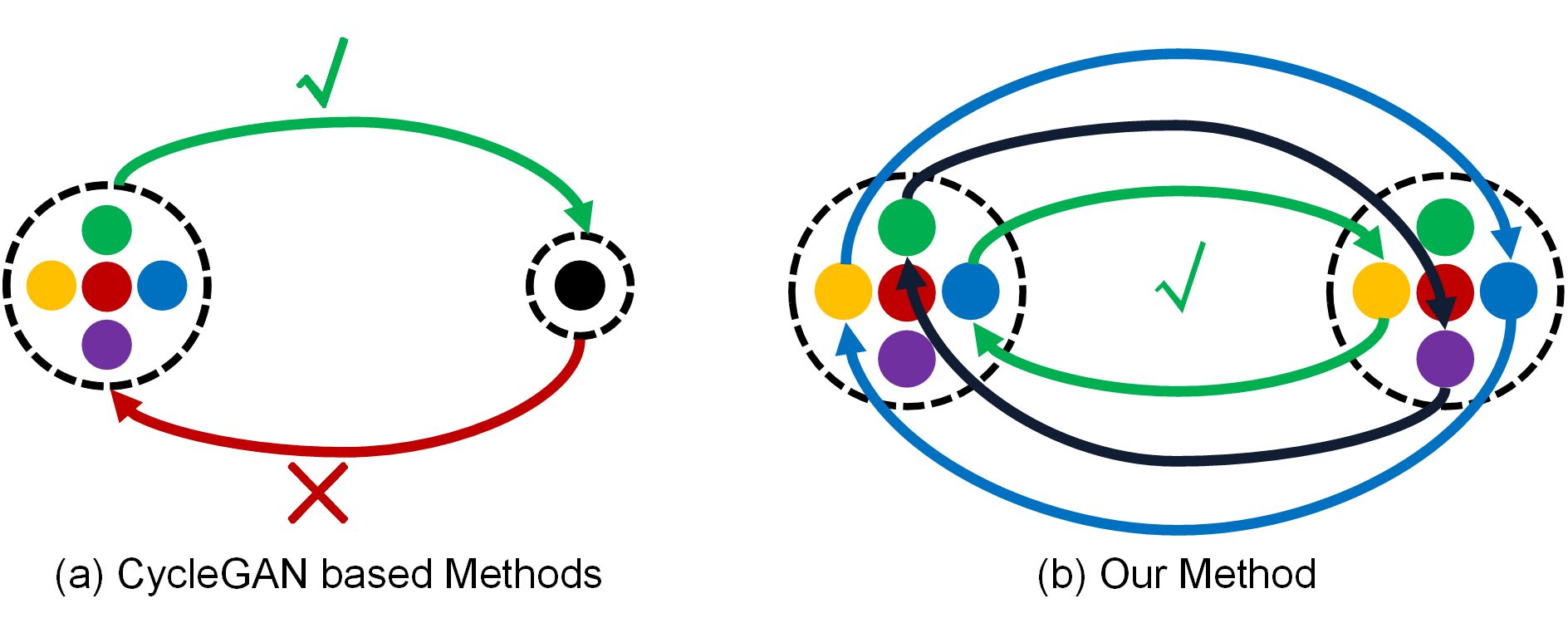}
	\caption{Mapping of different training frameworks. (a) Mapping of the CycleGAN-based method. Since it is not possible to determine where a single domain maps to, the mapping from a single domain to multiple domains is uncertain. (b) Mapping of StainPresetNet. In our training framework, the mapping relationship is clear and stable.}
	\label{fig2}
\end{figure}

\begin{figure*}[!ht]
	\centering
	\includegraphics[width=100 mm]{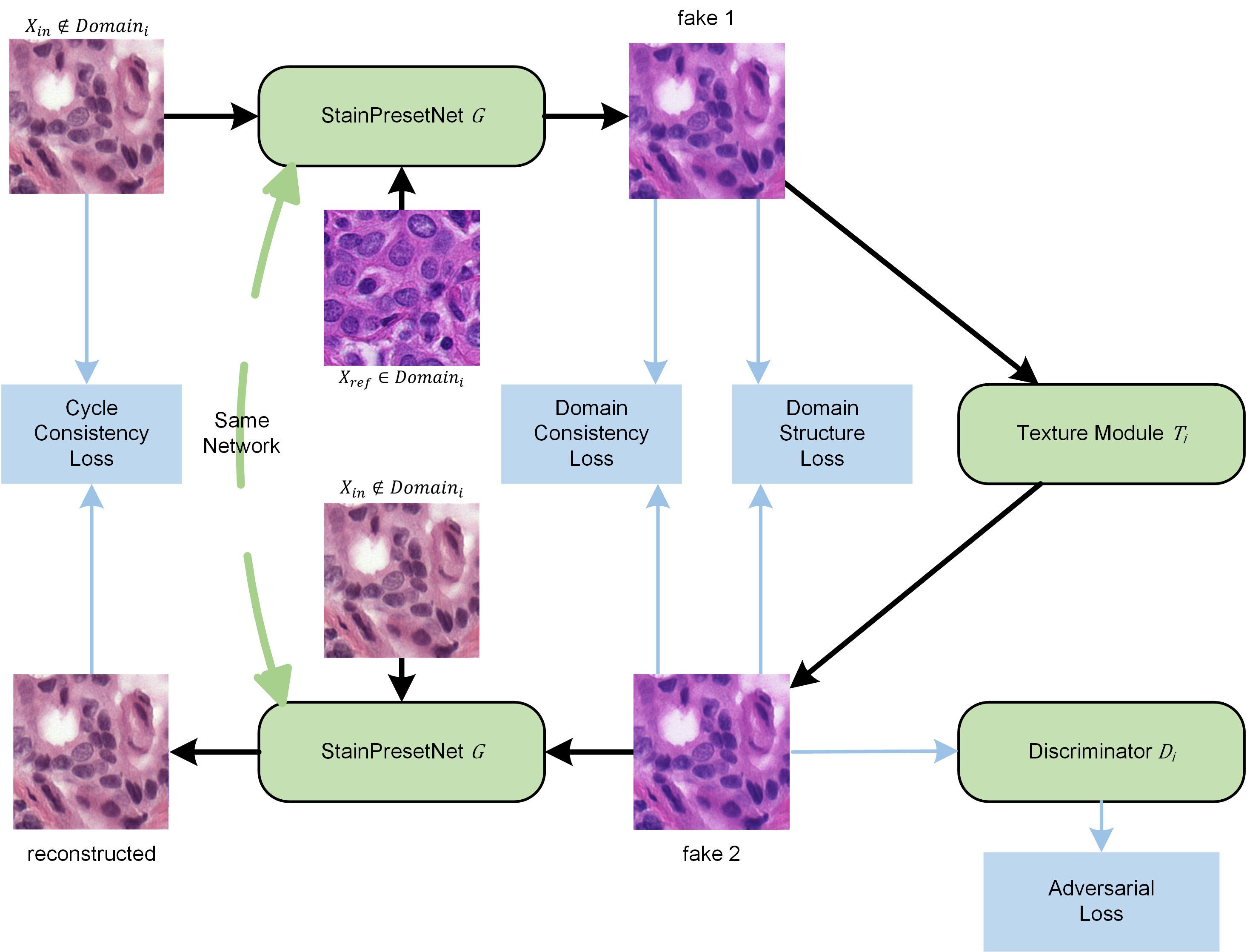}
	\caption{The training framework of StainPresetNet. A real imgae $x_{in} \notin Domain_i, i=1,2,...,N$ is mapped to $Domain_i$ under the reference of the real image $x_{ref} \in Domain_i$, the texture module $T_i \in Domain_i$ adds some details, and finally the real imgae $x_{in} \notin Domain_i$ is reconstructed under the reference of itself. $N$ is the number of domains.}
	\label{fig3}
\end{figure*}

Among stain normalization methods, CycleGAN is the mainstream training framework. However, CycleGAN is designed for one-to-one transformation tasks and cannot be applied to multi-to-one and multi-to-multi transformation tasks. The training framework of CycleGAN consists of two transformations: from domain A to domain B and from domain B to domain A. When the CycleGAN training framework is applied to multi-to-one domain transformation, the transformation from many to one is fine, but the transformation from one to many is uncertain as shown in Fig. \ref{fig2} (a). The uncertainty of the transformation makes the network training process ambiguous and confusing, resulting in reduced performance.

To solve this problem, a new training framework is proposed in this study, in which each domain is equipped with an independent texture module and discriminator, and the texture module and discriminator to be used are determined according to the domain of the reference image. The transformation of StainPresetNet in the multi-to-multi color normalization task is shown in Figure \ref{fig3} (b). For each pair of source image and reference image, the training framework of StainPresetNet performs the transformation between the two domains, with a clear mapping direction and more stable training.

The proposed training framework is shown in Fig. \ref{fig3}, which contains a StainPresetNet as the generator ($G$), $N$ texture modules ($T_i, i=1,2,3,...,N$) and $N$ discriminators ($D_i, i=1,2,3,...,N$), where $N$ is the number of domains in the training dataset. $G$ can transform the input source image $x_{in}$ to the reference image's under the reference of the reference image $x_{ref}$. The texture module $T_i$ is used to add some texture details belonging to $Domain_i$  to the output image of $G$, so as to deceive the discriminator $D_i$. The discriminator $D_i$ is used to distinguish the difference between the real reference image $x_{ref}$ and the output image of the texture module $T_i$. Next, the output image of the texture module $T_i$ will be transformed to the domain of $x_{in}$ under the reference of $x_{in}$ by $G$.

The proposed framework includes five loss functions: adversarial loss, cycle consistency loss, domain consistency loss, domain structure loss, and identity loss. The overall objective function can be expressed as:

\begin{equation}
	\begin{split}
		\mathcal { L } = &\sum\limits_{i=1}^{N}  \Big{[}\mathcal {L}_{GAN}(G,T_i,D_i)+ \lambda_{cyc} \mathcal {L}_{cyc}(G,T_i) \\&+ \lambda_{dom} \mathcal {L}_{dom}(G,T_i) + \lambda_{struct} \mathcal {L}_{struct}(G,T_i) \\&+ \lambda_{ide} \mathcal{L}_{ide}(G,T_i) \Big{]}
	\end{split}
\end{equation}
where, $\mathcal{L}_{GAN}(G,T_i)$ is the adversarial loss, $\mathcal {L}_{cyc}(G,T_i)$ is the cycle consistency loss, $\mathcal {L}_{dom}(G,T_i)$ is the domain consistency loss, $\mathcal {L}_{struct}(G,T_i)$ is the domain structure loss, $\mathcal{L}_{ide}(G,T_i)$ is the identity loss, $N$ is the number of domains in the dataset, $\lambda_{cyc}$, $\lambda_{dom}$, $\lambda_{struct}$ and $\lambda_{ide}$ are the weight coefficients of the cycle consistency loss, the domain consistency loss, the domain structure loss and the identity loss, respectively.

The adversarial loss is aimed to ensure that the output of the texture module $T_i$ have the same distribution as the real reference image $x_{ref}$. The objective function is defined as: 
\begin{equation} 
	\begin{split} 
		&\mathcal {L}_{GAN}(G,T_i,D_i)=\\&\mathbb{E}_{ x_{ref}\sim p_{data}(x_{ref})} [\Vert D_i(x_{ref})\Vert_2 ] \\&+ \mathbb{E}_{ x_{in}\sim p_{data}(x_{in})} [\Vert 1-D_i(T_i(G(x_{in}|x_{ref})))\Vert_2 ]\\&
		 {\rm where}\\&
		 x_{ref}\in {\rm Domain}_i, x_{in} \notin {\rm Domain}_i
		 \\& \Vert\cdot\Vert_2 \text{ is the } \ell_{2} \text{ norm}
		 \\& x_{in} \text{ is the input image}
		 \\& x_{ref} \text{ is the reference image}
		 \\
	\end{split}
\end{equation}

The cycle consistency loss constrains the consistency of image content by reconstructing the transformed image into the original image. The objective function can be expressed as:
\begin{equation}
	\begin{split}
		&\mathcal {L}_{cyc}(G,T_i)=   \\& \mathbb{E}_{ x_{in}\sim p_{data}(x_{in})} [\Vert x_{in}-G(T_i(G(x_{in}|x_{ref}))|x_{in})\Vert_1 ]\\
		&{\rm where} \\& x_{ref}\in {\rm Domain}_i, x_{in} \notin {\rm Domain}_i
		\\& \Vert\cdot\Vert_1 \text{ is the } \ell_{1} \text{ norm}
	\end{split}
\end{equation}

The domain consistency loss is designed that the output image of the generator $G$ and the output image of the texture module $T_i$ are as consistent as possible. The domain consistency loss uses the L1 distance to constrain the difference between the generator output image $G(x_{in}|x_{ref})$ and the texture module output image $T_i(G(x_{in}|x_{ref}))$. The objective function can be expressed as:
\begin{equation}
	\begin{split}
		&\mathcal {L}_{dom}(G,T_i)= \\&  \mathbb{E}_{ x_{in}\sim p_{data}(x_{in})} [\Vert G(x_{in}|x_{ref})-T_i(G(x_{in}|x_{ref}))\Vert_1 ]\\
		&{\rm where} \\
		& x_{ref}\in {\rm Domain}_i, x_{in} \notin {\rm Domain}_i
	\end{split}
\end{equation}

The domain structure loss is designed that the output image of the generator $G$ and the output image of the texture module $T_i$ are as consistent as possible in image structure. The domain structure loss uses SSIM to constrain the difference between the generator output image $G(x_{in}|x_{ref})$ and the texture module output image $T_i(G(x_{in}|x_{ref}))$. The objective function can be expressed as:
\begin{equation}
	\begin{split}
		&\mathcal {L}_{dom}(G,T_i)= \\&  \mathbb{E}_{ x_{in}\sim p_{data}(x_{in})} [
		1-SSIM(G(x_{in}|x_{ref}),T_i(G(x_{in}|x_{ref})) ]\\
		&{\rm where} \\
		& x_{ref}\in {\rm Domain}_i, x_{in} \notin {\rm Domain}_i
	\end{split}
\end{equation}

The identity loss is designed that when the input image comes from the domain of the reference image $x_{ref}$, the generator can maintain the identity mapping without making any changes to the image. Specifically, in this study, when the input of $G$ and $T_i$ is the reference image $x_{ref}$, $G$ and $T_i$ should not make any changes to the input. The objective function can be expressed as:
\begin{equation}
	\begin{split}
		\mathcal {L}_{ide}(G,T_i)=&\mathbb{E}_{ x_{ref}\sim p_{data}(x_{ref})} [\Vert x_{ref}-G(x_{ref}|x_{ref})\Vert_1 ] \\
		&+\mathbb{E}_{ x_{ref}\sim p_{data}(x_{ref})} [\Vert x_{ref}-T_i(x_{ref})\Vert_1 ]\\
		&{\rm where} \\ &x_{ref}\in {\rm Domain}_i, x_{in} \notin {\rm Domain}_i
	\end{split}
\end{equation}

In our implementation, the network structure of the texture module and discriminator use the generator and discriminator in ParamNet.

\section{Experiments and Results}
\subsection{Datasets}
In this study, we used three different datasets to verify the performance of different methods, including one public datasets and two private datasets. This study was approved by the Ethics Committee of Tongji Medical College, Huazhong University of Science and Technology. The related descriptions are below:
\subsubsection{Aligned Cytopathology Dataset}
In this dataset, the same slides (Thinprep cytologic test slides from Hubei Maternal and Child Health Hospital) were scanned using three different scanners. The first scanner was equipped with a 20x objective lens with a pixel size of 0.2930 um. The second scanner called was equipped with a 20x objective lens with a pixel size of 0.2436 um. The third scanner was equipped with a 40x objective lens with a pixel size of 0.1803 um. The images captured by these three scanners had three different styles, namely S1, S2, and S3. In this dataset, the images from different scanners are carefully aligned. There are 3,028 image pairs in the training set and 1,472 image pairs in the test set. The aligned cytopathology dataset was used to test the performance of the classifier with and without normalization

\begin{table}[!h]
	\centering
	\caption{The number of image patches on the cytopathology and histopathology classification dataset. The classification set was used to train or test the classifier with and without normalization. The normalization set was used to train stain normalization methods. }
	\label{table2}
	\setlength{\tabcolsep}{2pt}
	\begin{tabular}{ccc|ccc}
		\hline
		Styles &Classification&Normalization&Centers &Classification& Normalization\\
		\hline
		D1&244,341&2,500&Uni16&40,000&2,500\\
		D2&10,000&2,500&C1&13,613&2,500\\
		D3&10,000&2,500&C2&12,355&2,500\\
		D4&10,000&2,500&C3&14,076&2,500\\
		D5&10,000&2,500&C5&9,696&2,500\\
		&&&C5&9,696&2,500\\
		\hline
	\end{tabular}
\end{table}

\subsubsection{Cytopathology Classification Dataset}
The cytopathology classification dataset from ParamNet was used to verify the performance of the stain normalization algorithm on a multi-domain cytopathology classification diagnostic task. The slides in the cytopathology classification dataset are from Hubei Provincial Maternal and Child Health Hospital and Tongji Hospital. There are five different styles (D1-D5), among which D1-D4 come from Hubei Maternal and Child Health Hospital, and D5 comes from Tongji Hospital. In this dataset, the image patches that contain abnormal cells are labeled as abnormal patches, and the image patches that do not contain any abnormal cells are labeled as normal patches. There are 244,341 image patches in D1 and 40,000 image patches in D2-D5 (10,000 image patches in each style) as shown in Table \ref{table2}. We randomly picked 2,500 image patches in each style as the normalization set for training the stain normalization algorithm. For the classifier, we used D1 to train the classifier, and used D2-D5 to test the performance of the classifier with and without normalization.

\subsubsection{Histopathology Classification Dataset}
The histopathology classification datasets used the publicly available Camelyon16 \cite{camelyon16} (399 whole slide images from two centers) and Camelyon17 \cite{camelyon17} datasets (1,000 whole slide images from five centers). In our experiments, 100 WSIs from University Medical Center Utrecht (Uni16) in Camelyon16 training part were used to extract the training patches, and 500 WSIs from the five centers (C1-C5) in Camelyon17 training part were used to extract the test patches. In this dataset, the image patches that contain tumor cells are labeled as abnormal patches, and the image patches that do not contain any tumor cells are labeled as normal patches. There are 40,000 image patches from Uni16 in the training set of this dataset. And there are 58,975 image patches in the test set of this dataset as shown in Table \ref{table2}. We randomly picked 2,500 image patches in each center as the normalization set for training stain normalization algorithms. For the classifier, we used the training set to train the classifier, and used the test set to test the performance of the classifier with and without normalization.

\subsection{Evaluation Metrics}
In order to evaluate different methods fairly and effectively, we considered in four aspects: similarity with target image, preservation of source image information, computational efficiency, and classifier accuracy.

In order to effectively measure the similarity to the target image and preserve the information of the source image, we used two similarity matrices, Structural Similarity index (SSIM) and Peak Signal-to-Noise Ratio (PSNR). The similarity between the normalized image and the target image was evaluated using SSIM and PSNR, denoted as SSIM-T and PSNR-T. The preservation of source image information was evaluated using SSIM between the normalized  image and the source image, denoted as SSIM-S. SSIM-T and PSNR-T were calculated using raw RGB values. SSIM-S was calculated using grayscale images. 

In order to evaluate the computational efficiency of different methods, the number of frames per second (FPS) was calculated on the system with 6-core Intel(R) Core (TM) i7-6850K CPU and NVidia GeForce GTX 1080Ti. The input and output (IO) time was not included. The FPS results of different methods are shown in Table \ref{table4} when the input image size is 512×512 pixels.

The accuracy was used to evaluate the classifier performance. The statistic results of accuracy on the cytopathology and histopathology classification datasets are shown in Tables \ref{table5} and \ref{table6}, as mean ± standard deviation.

\subsection{Implementation}

For conventional methods (Reinhard \cite{reinhard2001color}, Macenko \cite{macenko2009method} and Vahadane \cite{vahadane2016structure}, one professionally selected image was used as the reference image. For the GAN based methods (StainGAN \cite{shaban2019staingan}, StainNet \cite{kang2021stainnet} and ParamNet \cite{kang2023paramnet}), we used their default training settings.

For StainPresetNet, the proposed framework included five loss functions: adversarial loss, cycle consistency loss, domain consistency loss, domain structure loss and identity loss, and the weights of these loss functions were 1.0, 10.0, 10.0, 1.0, and 2.0 respectively. During training, we randomly transformed image resolutions (0.5$\sim$1.0) to stabilize learning. We used Adam to optimize the network, and the learning rate is linearly increased from 0 to 1e-4 in the first 1,000 iterations, and then linearly decreased to 0 until the end of training during 2000k iterations.

For the classifier, we used SqueezeNet \cite{squeezenet} pre-trained on ImageNet \cite{imagenet} as the classification network. The classifier was trained using cross entropy loss and Adam optimizer. The initial learning rate was set to 0.0002, and it reduced by a factor of 0.1 at the 40th and 50th epoch. The training was stopped at the 60th epoch, which was chosen experimentally. The experiment was repeated 10 times in order to enhance reliability.

\subsection{Results}
In this section, we compared our method with state-of-art normalization methods, including Reinhard, Macenko, Vahadane, StainGAN, StainNet, and ParamNet.

\subsubsection{Image Similarity Evaluation}

We compared our StainPresetNet with state-of-art normalization methods on the aligned cytopathology dataset. We reported the following results: (1) Visual comparison among different methods. (2) Quantitative comparison of similarities among different methods. (3) The impact of reference images. (4) Quantitative comparison of computational efficiency among different methods.
\begin{figure*}[!ht]
	\centering
	\includegraphics[width=132 mm]{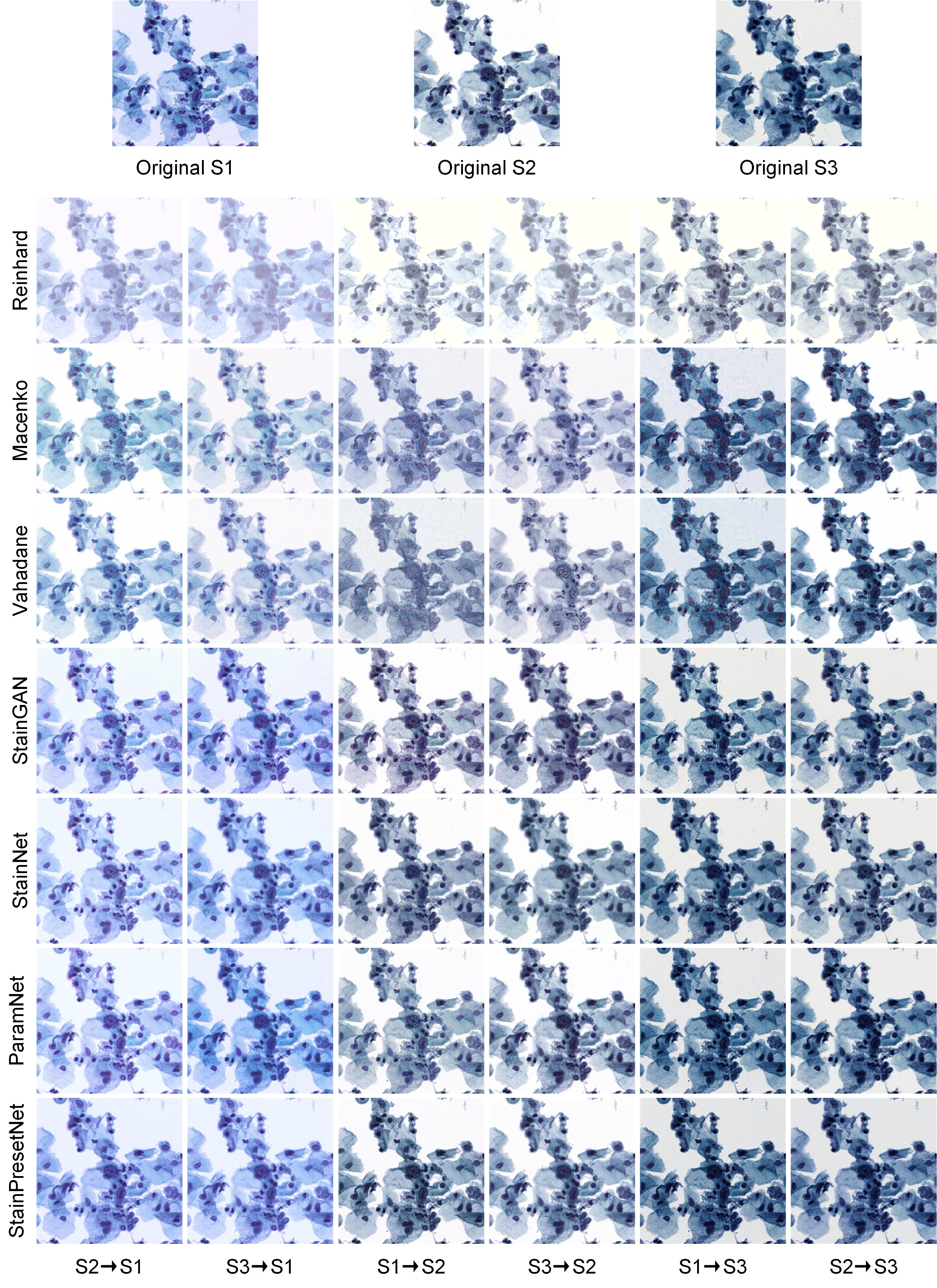}
	\caption{Visual comparison among different methods on the aligned cytopathology dataset. To achieve transformations among multiple styles, StainGAN, StainNet, and ParamNet were trained multiple times, but our StainPresetNet only was trained once.}
	\label{fig4}
\end{figure*}

\begin{table*}[!ht]
	\centering
	\caption{Different evaluation metrics are reported for various stain normalization methods on the aligned cytopathology dataset.To achieve transformations among multiple styles, StainGAN, StainNet, and ParamNet were trained multiple times, but our StainPresetNet only was trained once.}
	\label{table3}
	\setlength{\tabcolsep}{5pt}
	\begin{tabular}{cccc|ccc|ccc}
		\hline
		\multirow{2}{*} {Methods}&\multicolumn{3}{c}{To S1}& \multicolumn{3}{c}{To S2}&\multicolumn{3}{c}{To S3}\\
		& SSIM-T & PSNR-T & SSIM-S & SSIM-T & PSNR-T & SSIM-S & SSIM-T & PSNR-T & SSIM-S  \\
		\hline
		Reinhard    & 0.901±0.031 &  24.0±3.6   & 0.920±0.049 & 0.908±0.033 &  24.5±5.1   & 0.931±0.038 & 0.844±0.032 &  23.5±4.8   & 0.970±0.019 \\
		Macenko     & 0.896±0.030 &  23.6±2.6   & 0.946±0.035 & 0.884±0.036 &  22.8±2.4   & 0.956±0.029 & 0.830±0.039 &  22.5±2.1   & 0.933±0.032 \\
		Vahadane    & 0.896±0.034 &  23.8±2.8   & 0.945±0.036 & 0.888±0.034 &  23.0±2.7   & 0.944±0.036 & 0.831±0.043 &  22.6±2.1   & 0.939±0.038 \\
		StainGAN    & 0.891±0.023 &  28.4±2.5   & 0.910±0.021 & 0.901±0.026 &  28.5±2.8   & 0.915±0.017 & 0.830±0.021 &  27.6±2.2   & 0.911±0.022 \\
		StainNet    & 0.916±0.022 &  28.4±2.7   & 0.925±0.021 & 0.924±0.023 &  29.4±3.1   & 0.941±0.011 & 0.870±0.022 &  27.6±2.3   & 0.959±0.013 \\
		ParamNet    & 0.919±0.020 &  27.9±2.8   & \textbf{0.960±0.014} & 0.922±0.023 &  28.8±3.2   & \textbf{0.972±0.010} & 0.871±0.022 &  29.3±2.8   & \textbf{0.970±0.014} \\
		\textbf{StainPresetNet} & \textbf{0.922±0.020} &  \textbf{30.1±2.5}   & 0.952±0.012 & \textbf{0.929±0.022} &  \textbf{30.4±3.3}   & 0.950±0.007 & \textbf{0.874±0.021} &  \textbf{29.8±2.9}   & 0.962±0.013\\
		\hline
	\end{tabular}
\end{table*}

\begin{figure*}[htbp]    
		\subfloat[S2 and S3 to S1]   
		{
			\label{subfig1}\includegraphics[width=0.3 \textwidth]{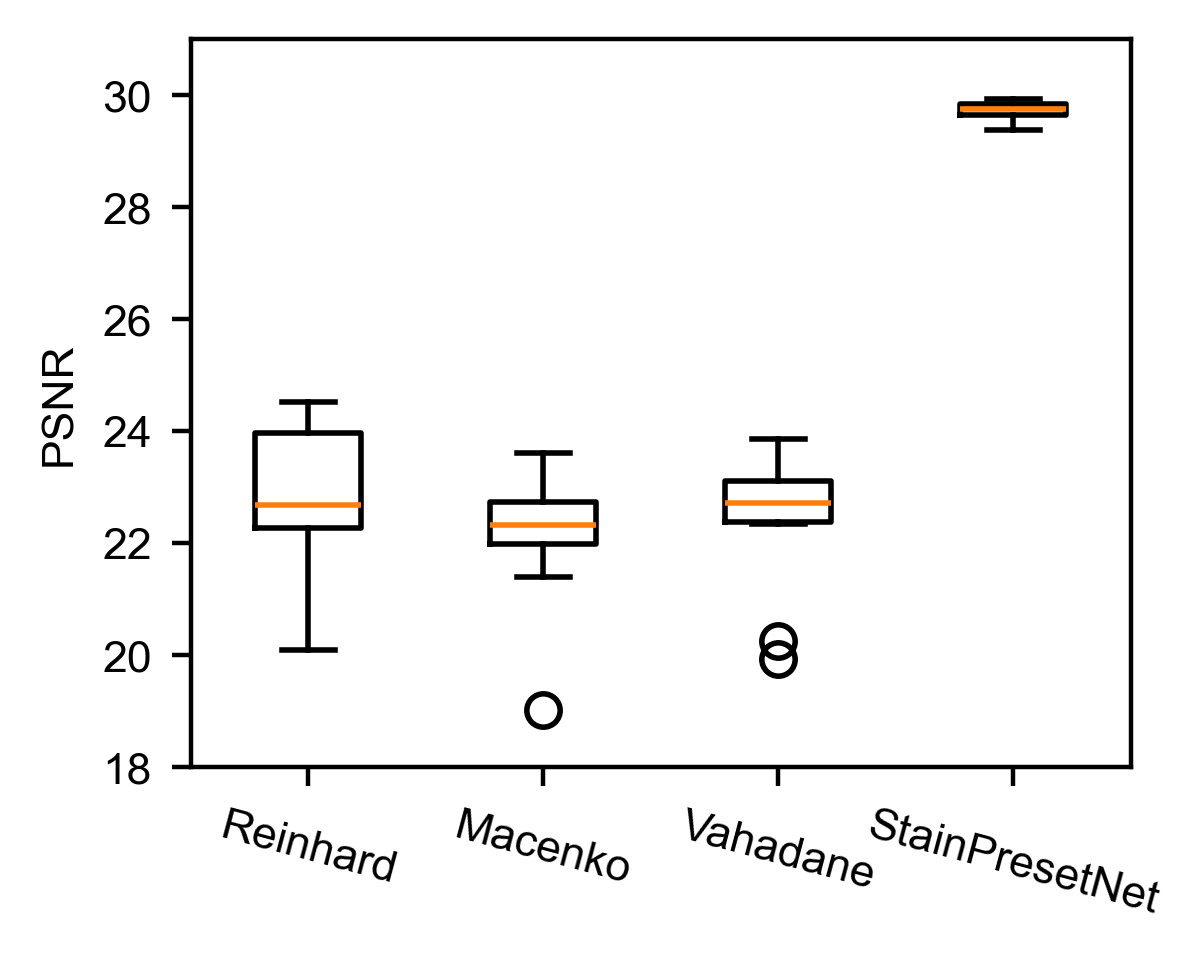}
		}
		\subfloat[S1 and S3 to S2]
		{
			\label{subfig2}\includegraphics[width=0.3 \textwidth]{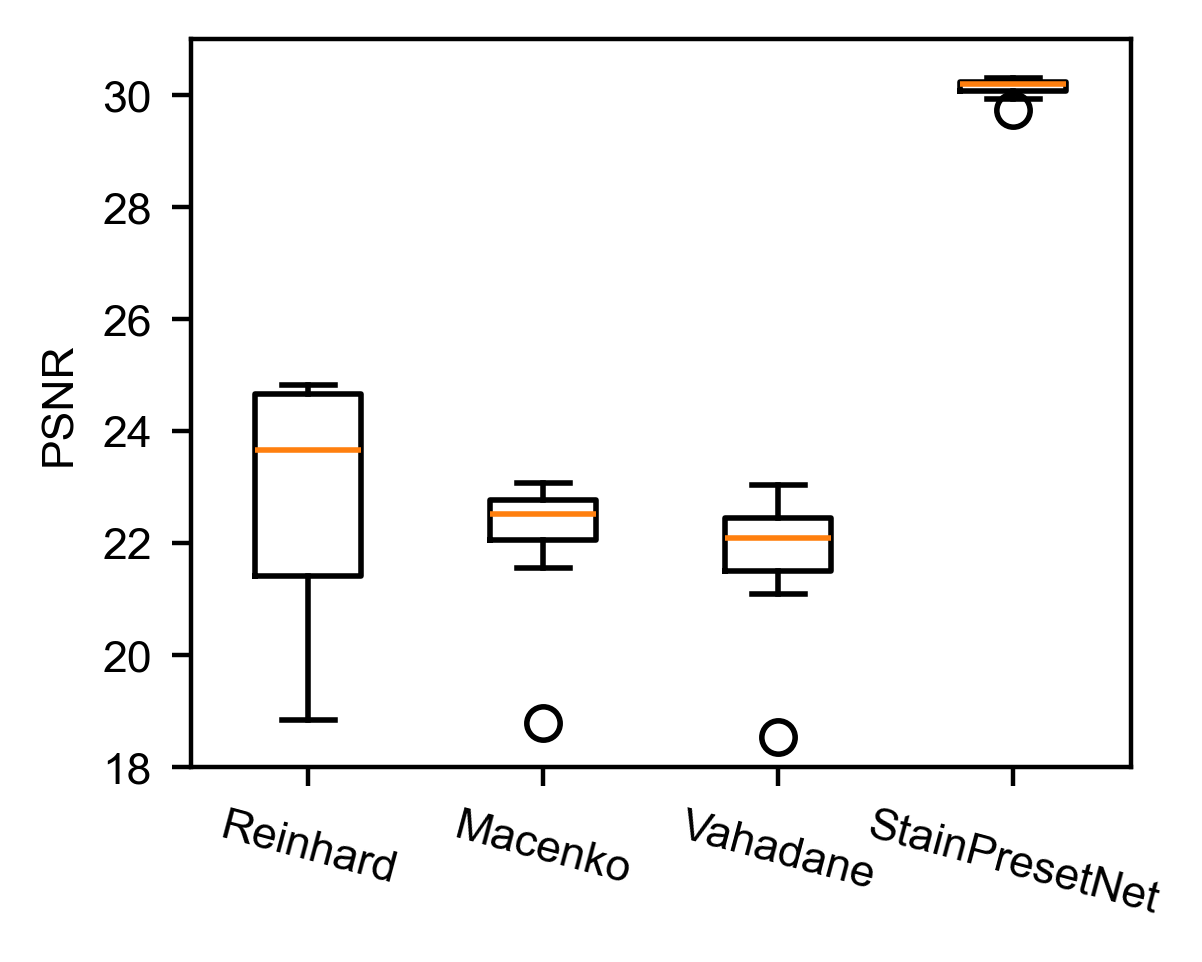}
		}
		\subfloat[S1 and S2 to S3]
		{
			\label{subfig3}\includegraphics[width=0.3 \textwidth]{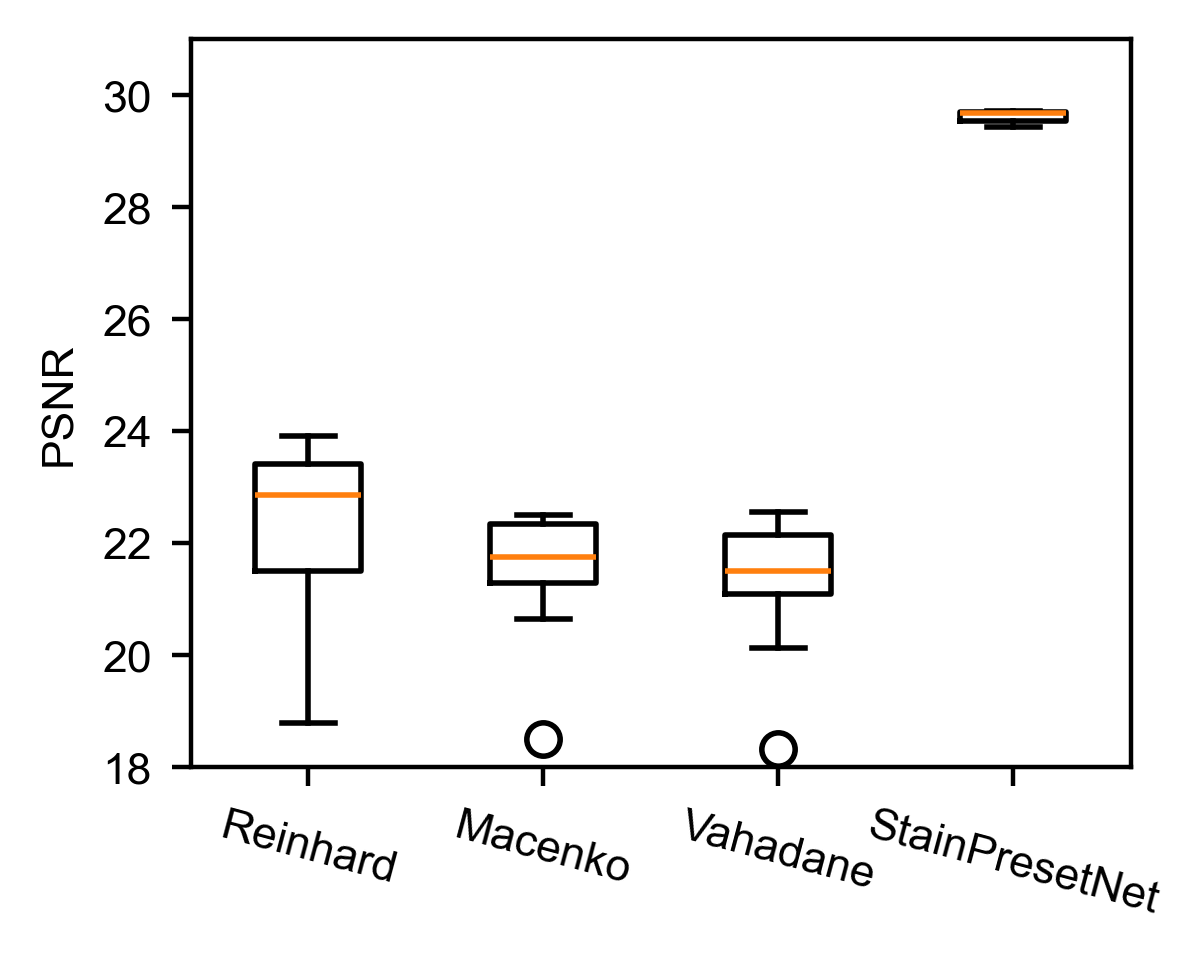}
		}
		\caption{The impact of reference images on the performance of different methods. Ten images of each style were randomly selected as reference images on the aligned cytopathology dataset.}    
		\label{fig5}          
\end{figure*}

\begin{table*}[!h]
	\footnotesize 
	\centering
	\caption{Quantitative comparison of computational efficiency among different methods. The input and output (IO) time is not included. The input image size is 512×512 pixels.}
	\label{table4}
	\begin{tabular}{cccccccc}
		\hline
		Methods& Reinhard& Macenko& Vahadane& StainGAN& StainNet& ParamNet& \textbf{StainPresetNet}\\
		\hline
		FPS& 54.8& 4.0& 0.5&  19.6& 881.8& \textbf{1,605.2}&1,601.5 \\
		\hline
	\end{tabular}
\end{table*}

\textbf{Visual comparison}

Firstly, we compared our StainPresetNet with other stain normalization methods in the visual appearance. The visual comparison among different methods is shown in Fig. \ref{fig4}. The aligned cytopathology dataset contains three different styles S1, S2, and S3, so we performed normalization in three different directions. The normalized images by the conventional methods (Reinhard, Macenko and Vahadane) has a large gap with the corresponding real target images in brightness, contrast and color. For StainGAN, StainNet, and ParamNet, these methods cannot change the direction of normalization during inference. In order to achieve transformations among multiple styles, these methods were trained multiple times. Even if each style is learned, these methods still cannot achieve the best performance. Especially on S3→S1, the images normalized by these methods were darker than the real S1 images. Our StainPresetNet only needs to be trained once to achieve the transformation between multiple styles, and it is visually closest to the corresponding target image.

\textbf{Quantitative comparison of similarities}

Secondly, we quantitatively compared our StainPresetNet with other stain normalization methods in the similarity. The results on the aligned cytopathology dataset are shown in Table \ref{table3}, which contains quantitative evaluations in three different normalization directions. The conventional methods (Reinhard, Macenko, and Vahadane) have low SSIM-T and PSNR-T, which means that there is a large difference between the images normalized by the conventional methods and the target images. Our StainPresetNet have highest SSIM-T and PSNR-T in all normalization directions, which means the images normalized by our StainPresetNet is closest to the corresponding target image. In Table \ref{table3}, among the deep learning-based stain normalization methods, StainPresetNet has a relatively high SSIM-S, and ParamNet has the highest SSIM-S. StainGAN and StainNet do not obtain the best matrices in any style, even though they learn each style.

\textbf{The impact of reference images}

Conventional methods (Reinhard, Macenko, and Vahadane) and our StainPresetNet extract color information from the reference image to determine the direction of color normalization. We verified the impact of reference images on the performance of different methods on the aligned cytopathology dataset. Specifically, for each style, we randomly selected an image as the reference image, and then used this reference image to normalize images of other styles and calculate the PSNR with the target image. This experiment was repeated 10 times, and the results are shown in Fig. \ref{fig5}. 

As shown in Fig. \ref{fig5}, the reference image has a great impact on the performance of the conventional methods  (Reinhard, Macenko, and Vahadane), which shows that the conventional methods is very dependent on tthe reference image. For our method, the reference image has little impact and the performance is significantly better than the conventional method, which shows that our method can extract accurate color distribution information from a single reference image.

\textbf{Computational efficiency}

In this section, the computational efficiency of the proposed method is quantitatively evaluated compared with other color normalization methods. The FPS of different methods is tested when the input image size is 512×512, and the input and output time of the data is not included. The results are calculated on a system equipped with an Intel i7-6850K CPU and an NVIDIA GeForce GTX 1080Ti graphics card and are shown in Table \ref{table4}. As shown in Table \ref{table4}, the proposed StainPresetNet (1,601.5 FPS) is almost as fast as ParamNet (1,605.2 FPS). This is because the color distribution feature vector of the reference image is pre-computed and does not take up computing time. Compared with ParamNet, the additional computation comes from the fusion layer. In our StainPresetNet, the fusion layer has only two fully connected layers, one of which has a 118-dimensional feature vector as input and output, and the other is a fully connected layer with a 118-dimensional input and a 59-dimensional output. Therefore, the fusion layer does not introduce too much computation, and the StainPresetNet proposed in this paper is almost as fast as ParamNet.

\subsection{Patch-Level Classification Evaluation }
\begin{table*}[!ht]
	\centering
	\caption{The accuracy for various stain normalization methods on the cytopathology classification dataset.}
	\label{table5}
	\begin{tabular}{cccccc}
		\hline
		Accuracy& D2& D3& D4& D5& Average\\
		\hline
		Original& 0.853±0.031& 0.915±0.022& 0.873±0.029& 0.688±0.033& 0.832±0.029\\
		Reinhard& 0.862±0.015& 0.867±0.025& 0.796±0.026& 0.728±0.025& 0.813±0.023\\
		Macenko& 0.929±0.014& 0.937±0.023& 0.915±0.010& 0.818±0.034& 0.900±0.020\\
		Vahadane& 0.905±0.022& 0.940±0.031& 0.907±0.019& 0.745±0.037& 0.875±0.026\\
		StainGAN& 0.918±0.007& 0.908±0.011& 0.867±0.006& 0.799±0.018& 0.873±0.011\\
		StainNet& 0.928±0.034& 0.963±0.014& 0.924±0.032& 0.820±0.059& 0.909±0.035\\
		ParamNet& 0.968±0.013& 0.965±0.017& 0.943±0.006& 0.867±0.039& 0.936±0.019\\
		\textbf{StainPresetNet}& \textbf{0.971±0.010}& \textbf{0.976±0.011}& \textbf{0.950±0.003}&\textbf{ 0.907±0.035}&\textbf{0.951±0.015}\\
		\hline
	\end{tabular}
\end{table*}

\begin{table*}[!ht]
	\centering
	\caption{The accuracy for various stain normalization methods on the histopathology classification dataset. C3 and Uni16 are from the same medical center, but at different times.}
	\label{table6}
	\setlength{\tabcolsep}{3pt}
	\begin{tabular}{ccccccc}
		\hline
		Accuracy&C1&C2& C3&C4&C5&Average\\
		\hline
		Original& 0.500±0.016& 0.561±0.030& 0.907±0.003& 0.528±0.014& 0.765±0.041& 0.652±0.021\\
		Reinhard& 0.871±0.014& 0.855±0.008& 0.815±0.006& 0.823±0.027& 0.808±0.015& 0.834±0.014\\
		Macenko& 0.814±0.009& 0.832±0.009& 0.816±0.007& 0.836±0.012& 0.789±0.010& 0.817±0.009\\
		Vahadane& 0.817±0.005& 0.826±0.005& 0.762±0.007& 0.890±0.004& 0.805±0.003& 0.820±0.005\\
		StainGAN& 0.898±0.004& 0.871±0.001& 0.921±0.003& 0.903±0.004& 0.881±0.005& 0.895±0.003\\
		StainNet& 0.898±0.006& 0.836±0.006& 0.895±0.005& 0.862±0.019& 0.877±0.005& 0.874±0.008\\
		ParamNet& 0.920±0.003& 0.857±0.007&\textbf{0.924±0.003}& 0.899±0.004& 0.884±0.003& 0.897±0.004\\
		\textbf{StainPresetNet}& \textbf{0.924±0.002}&\textbf{0.884±0.005}&0.919±0.003& \textbf{0.914±0.003}& \textbf{0.895±0.001}& \textbf{0.907±0.003}\\
		\hline
	\end{tabular}
\end{table*}

In this section, the proposed StainPresetNet and other stain normalization methods are evaluated on patch-level classification tasks on datasets from multiple sources. SqueezeNet pre-trained on ImageNet is used as the classifier because of its small size and relatively high accuracy. The accuracy of the classifier without normalization and with various normalization methods is shown in Table \ref{table5} and Table \ref{table6}.

The accuracy of the classifier without normalization and with various normalization methods on the cytopathology classification dataset is shown in Table \ref{table5}.  As shown in Table \ref{table5}, the performance of the classifier on the original images of D2-D5 is poor, and stain normalization can effectively improve the performance of the classifier. The classifier performs better on images normalized by deep learning-based methods than conventional methods. Among deep learning-based methods, the classifier has the highest accuracy on the images normalized by StainPresetNet, with an average of 0.951. 

The accuracy without normalization and with various normalization methods on the histopathology classification dataset is shown in Table \ref{table6}. As shown in Table \ref{table6}, the classifier performs poorly on the original images of C1-C5. Using stain normalization can significantly improve the accuracy of the classifier, among which the deep learning-based methods are significantly better than the conventional methods. In Table \ref{table6}, our StainPresetNet has the best average accuracy of 0.907. The images in C3 and Uni16 are from the same medical center but at different times. For the classifier trained on Uni16, the accuracy is 0.907 on the original images of C3. And the accuracy can be increased to 0.919 on the normalized images by our StainPresetNet, which shows that our method can improve the accuracy of the classifier, even on images from the same center.

\subsection{WSI-Level Classification Evaluation}

\begin{table*}[!ht]
	\centering
	\caption{Comparison of different stain normalization methods on the histopathological WSIs from different medical centers. C3 and Uni16 are from the same medical center, but at different times.}
	\begin{tabular}{cccccccc}
		\hline
		Centers &     Methods     & Precision & Sensitivity & Specificity & Accuracy& F1 score  & AUC \\
		\hline
		C1&	Original&0.000&0.000&\textbf{1.000}&0.998&0.000&0.500\\
		C1	&StainGAN&0.300&0.840&0.995&0.995&0.442&0.932\\
		C1&	StainNet&\textbf{0.709}&0.822&0.999&\textbf{0.999}&\textbf{0.761}&0.905\\
		C1&	ParamNet&0.372&\textbf{0.858}&0.997&0.996&0.519&\textbf{0.929}\\
		C1	&StainPresetNet&0.591&0.842&0.999&0.998&0.695&0.917\\
		\hline
		C2&	Original&0.000&0.000&\textbf{1.000}&\textbf{0.998}&0.000&0.500\\
		C2&	StainGAN&0.050&\textbf{0.753}&0.977&0.977&0.094&\textbf{0.875}\\
		C2&	StainNet&0.111&0.477&0.994&0.993&0.180&0.747\\
		C2	&ParamNet&0.128&0.627&0.993&0.993&0.213&0.814\\
		C2&	StainPresetNet&\textbf{0.196}&0.668&0.996&0.995&\textbf{0.303}&0.834\\
		\hline
		C3&	Original&0.770&0.819&\textbf{0.999}&\textbf{0.998}&0.794&0.929\\
		C3&	StainGAN&0.767&\textbf{0.894}&0.998&\textbf{0.998}&0.825&\textbf{0.952}\\
		C3	&StainNet&0.630&0.863&0.997&0.996&0.728&0.936\\
		C3&	ParamNet&\textbf{0.832}&0.854&\textbf{0.999}&\textbf{0.998}&\textbf{0.843}&0.933\\
		C3&StainPresetNet&0.817&0.861&\textbf{0.999}&\textbf{0.998}&0.839&0.939\\
		\hline
		C4&	Original&0.007&0.000&\textbf{1.000}&0.994&0.000&0.500\\
		C4&	StainGAN&0.348&0.812&0.991&0.990&0.487&\textbf{0.932}\\
		C4&	StainNet&\textbf{0.793}&0.122&\textbf{1.000}&0.995&0.212&0.593\\
		C4&	ParamNet&0.672&0.838&0.998&\textbf{0.997}&0.746&0.925\\
		C4&	StainPresetNet&0.713&\textbf{0.861}&0.998&\textbf{0.997}&\textbf{0.780}&0.930\\
		\hline
		C5&	Original&0.465&0.874&0.982&0.980&0.607&0.953\\
		C5&	StainGAN&0.606&0.662&0.992&0.987&0.633&0.907\\
		C5&	StainNet&\textbf{0.663}&0.836&0.992&0.990&0.739&0.949\\
		C5&	ParamNet&0.552&0.913&0.987&0.985&0.688&0.977\\
		C5&	StainPresetNet&0.837&\textbf{0.935}&\textbf{0.997}&\textbf{0.996}&\textbf{0.883}&\textbf{0.983}\\
		\hline
		Average&Original&0.209&0.282&\textbf{0.997}&0.994&0.234&0.647\\
		Average&StainGAN&0.397&0.739&0.991&0.989&0.476&0.904\\
		Average&StainNet&0.565&0.562&\textbf{0.997}&0.994&0.492&0.801\\
		Average&ParamNet&0.495&0.773&0.995&0.993&0.580&0.901\\
		Average&StainPresetNet&\textbf{0.588}&\textbf{0.795}&\textbf{0.997}&\textbf{0.996}&\textbf{0.660}&\textbf{0.906}\\
		\hline
	\end{tabular}
	\label{tab5-4}
\end{table*}

In this section, we evaluate the proposed StainPresetNet and other stain normalization methods on the whole slide image level (WSI level) classification task on Camelyon16 and Camelyon17 datasets. The training dataset contained 60 negative whole slide images and 40 positive whole slide images from Camelyon16, University Medical Center Utrecht (Uni16). The testing dataset contained 50 positive whole slide images from Camelyon17 (10 whole slide images from each of the 5 centers) to test the performance of the classifier with and without normalization. The classifier predicts the tissue area of the whole slide image in a sliding window manner to obtain a probability heat map.


Quantitative evaluations on C1-C5 using different normalization methods are shown in Table \ref{tab5-4}. As can be seen from Table \ref{tab5-4}, the classifier performs very poorly on unnormalized WSIs, with an F1 score of 0.000 at C1, C2, and C4. Classifier performance can be significantly improved using normalization methods. Among the normalization methods in Table\ref{tab5-4}, StainGAN failed to obtain the best performance at any center, StainNet could obtain the highest F1 score on C1, ParamNet could obtain the best F1 score on C3, and StainPresetNet Best F1 scores were achieved on C2, C4 and C5. In addition, for the average metrics on the C1-C5, StainPresetNet has obtained all the best metrics, which shows that StainPresetNet is more stable and has best performance. In particular, the F1 score of StainPresetNet is 8.0\% higher than the second place.

\section{Discussion and Conclusion}

In this paper, we proposed a novel stain normalization method named StainPresetNet, which achieves accurate mapping relationships, high computational efficiency, structure preservation, and multi-domain applicability. StainPresetNet extracts color mapping relationships from entire datasets, normalizes on a pixel-by-pixel manner, and utilizes a preset reference image to guide the normalization direction, all while maintaining high computational efficiency. 

Our proposed StainPresetNet can achieve the state-of-art performance on multiple style mutual transformation tasks, multi-center image patch-level classification tasks and multi-center WSI-level classification tasks. Our StainPresetNet can process 1,601.5 512×512 images per second using a single 1080ti graphics card, which means that for a 100,000×100,000 whole slide image, it only takes about 25 seconds to process. These results prove that StainPresetNet has the ability to be applied in CAD systems.

\bibliographystyle{ieeetr}
\bibliography{refs-used.bib}
\end{document}